\documentclass[10pt, journal]{IEEEtran}
\usepackage{array}
\usepackage{graphicx}
\usepackage{amsmath}
\usepackage{amssymb}
\usepackage{bm}
\usepackage{booktabs}
\usepackage{pifont}
\usepackage{adjustbox}
\usepackage{multirow}
\usepackage{makecell}
\usepackage{url}
\usepackage{subfigure}

\usepackage{xcolor}

\newcolumntype{C}[1]{>{\centering\arraybackslash}p{#1}}

\makeatletter
\newcommand{\thickhline}{%
	\noalign {\ifnum 0=`}\fi \hrule height 1pt
	\futurelet \reserved@a \@xhline
}
\newcolumntype{"}{@{\vrule width 1pt}}
\makeatother

\ifCLASSINFOpdf
\else
\fi

\begin{document}
	
\title{Coordinate-Based Dual-Constrained Autoregressive Motion Generation}

\author{Kang~Ding, Hongsong~Wang, Jie~Gui, and~Liang~Wang, Fellow, IEEE 
	\IEEEcompsocitemizethanks{
		\IEEEcompsocthanksitem K.~Ding is with School of Cyber Science and Engineering, Southeast University, Nanjing 210096, China (220245800@seu.edu.cn).
		\IEEEcompsocthanksitem H.~Wang is with School of Computer Science and Engineering, Key Laboratory of New Generation Artificial Intelligence Technology and Its Interdisciplinary Applications, Ministry of Education, Southeast University, Nanjing 210096, China (hongsongwang@seu.edu.cn).
		\IEEEcompsocthanksitem J.~Gui is with School of Cyber Science and Engineering, Southeast University, also with Purple Mountain Laboratories, Nanjing 211111, China, and also with Engineering Research Center of Blockchain Application, Supervision And Management (Southeast University), Ministry of Education, Southeast University, Nanjing 210096, China (guijie@seu.edu.cn).
		\IEEEcompsocthanksitem L. Wang is with New Laboratory of Pattern Recognition (NLPR), also with State Key Laboratory of Multimodal Artificial Intelligence Systems (MAIS), Institute of Automation, Chinese Academy of Sciences (CASIA), and also with School of Artificial Intelligence, University of Chinese Academy of Sciences (wangliang@nlpr.ia.ac.cn).
	}
}

\maketitle

\begin{abstract}
Text-to-motion generation has attracted increasing attention in the research community recently, with potential applications in animation, virtual reality, robotics, and human–computer interaction. Diffusion and autoregressive models are two popular and parallel research directions for text-to-motion generation. However, diffusion models often suffer from error amplification during noise prediction, while autoregressive models exhibit mode collapse due to motion discretization. To address these limitations, we propose a flexible, high-fidelity, and semantically faithful text-to-motion framework, named Coordinate-based Dual-constrained Autoregressive Motion Generation (CDAMD). With motion coordinates as input, CDAMD follows the autoregressive paradigm and leverages diffusion-inspired multi-layer perceptrons to enhance the fidelity of predicted motions. Furthermore, a Dual-Constrained Causal Mask is introduced to guide autoregressive generation, where motion tokens act as priors and are concatenated with textual encodings. Since there is limited work on coordinate-based motion synthesis, we establish new benchmarks for both text-to-motion generation and motion editing. Experimental results demonstrate that our approach achieves state-of-the-art performance in terms of both fidelity and semantic consistency on these benchmarks. Code is available at: \url{https://github.com/fly-dk/CDAMD}
\end{abstract}

\begin{IEEEkeywords}
Text-to-motion generation, motion synthesis, autoregressive model, motion diffusion
\end{IEEEkeywords}

%
\IEEEpeerreviewmaketitle
\section{Introduction}\label{sec:intro}
\IEEEPARstart{H}{uman} motion synthesis~\cite{zhu2023human} aims to generate realistic and coherent human movements under diverse conditions, such as textual descriptions, audio signals, or contextual constraints. 
Among various modalities, textual descriptions offer the most flexible and expressive interface, and text-driven motion synthesis has emerged as a promising direction~\cite{khani2025motion,sahili2025text}.

Inspired by the success of latent diffusion in text-to-image generation~\cite{rombach2022high,DBLP:conf/iccv/ZhangRA23}, diffusion models have recently become a popular framework for text-to-motion synthesis~\cite{chen2023executing,tan2024sopo,weng2026realign,wang2026temporal}. Most existing diffusion-based methods typically rely on an iterative denoising process, adopt mixed motion representations encompassing joint positions and kinematic rotations. However, these models suffer from a dimensional distribution mismatch as the hybrid motion representation fails to align with the standard normal distribution assumption. 
In addition, noise prediction in diffusion models faces error accumulation, as the standard deviation ratio from normalization exacerbates noise errors~\cite{meng2025rethinking}.

\begin{figure}[tbp]    
	\centering    
	\includegraphics[width=0.48\textwidth]{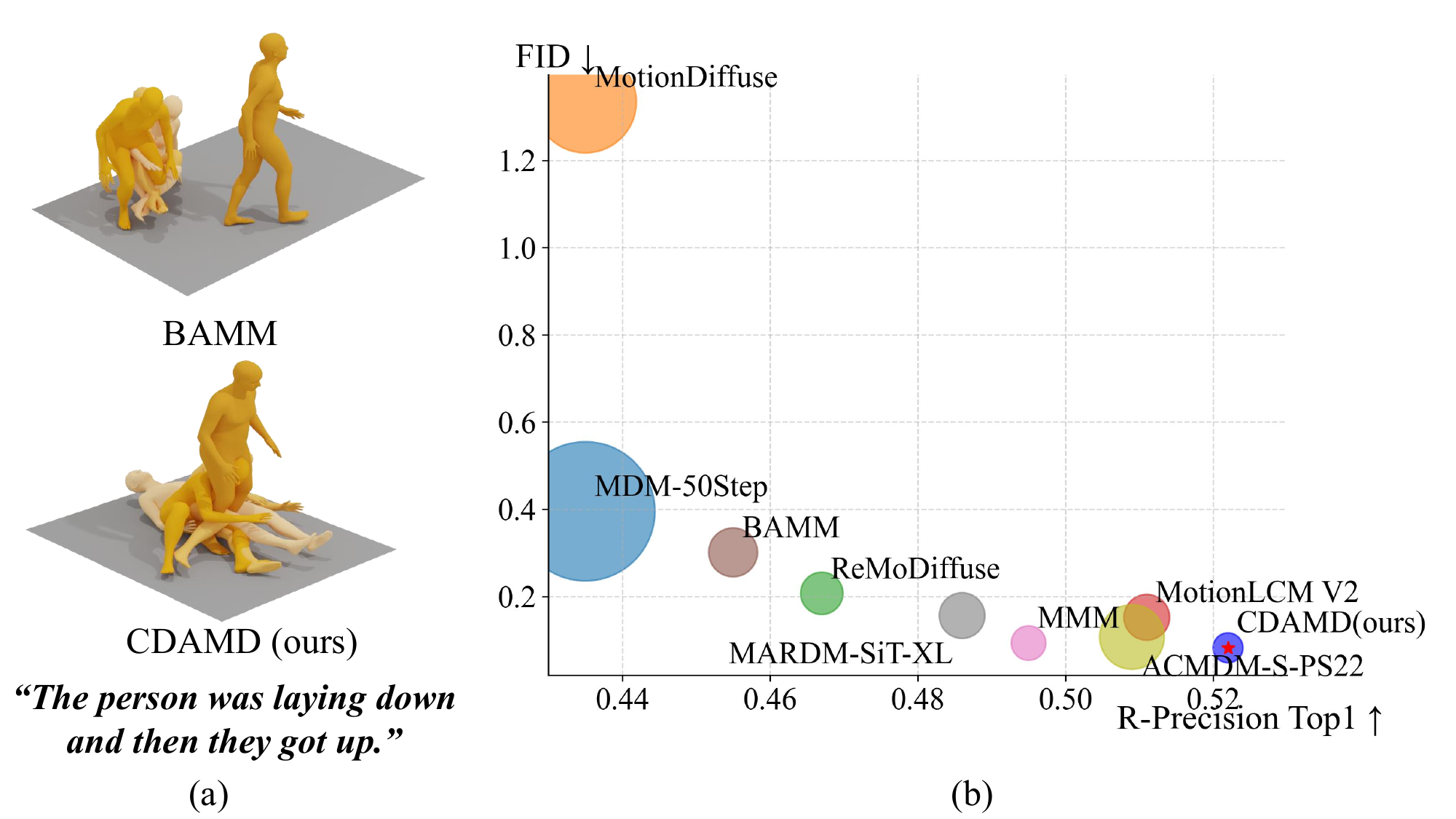}    
	\caption{Comparison of our approach with existing methods: (a) The existing autoregressive model (BAMM~\cite{pinyoanuntapong2024bamm}) fails to generate text-aligned motion for this example text prompt; (b) Top-1 R-Precision and FID results on the HumanML3D (bubble areas represent the 95\% confidence interval of the FID). Our approach achieves excellent performance in both FID and R-Precision.}    
	\label{fig:intro}
\end{figure}

Autoregressive motion models~\cite{pinyoanuntapong2024bamm,zhang2023generating,zhang2024motiongpt,hwangsnapmogen,tuautoregressive} encode human motion into discrete tokens and leverage autoregressive architectures to predict subsequent tokens. These models effectively capture long-range temporal dependencies and mitigate the error accumulations issues plaguing continuous diffusion models. 
For example, BAMM~\cite{pinyoanuntapong2024bamm} introduces a bidirectional causal mask, enabling next-token motion prediction to utilize both past tokens and future unmasked tokens. However, this bidirectional masking also leads to generated motion sequences that fail to faithfully follow the intended text semantics, as illustrated in Fig.~\ref{fig:intro}(a). 
In addition, autoregressive models often result mode collapse, as motion discretization inevitably causes the loss of fine-grained details.

Motion representation is also a crucial aspect of human motion synthesis, where the \textit{de facto} practice adopts a mixed representation of local-relative joint positions and kinematic-aware 6D rotations. This mixed representation introduces redundancy, and the rotation component does not align well with the conditional signals provided by users. For motion editing, users often specify the target joint position of a human motion sequence. Nevertheless, there exist only a few studies that use joint coordinates for motion generation~\cite{meng2025absolute}.

Following the autoregressive text-to-motion paradigm, we design a flexible, high-fidelity, and semantically faithful framework, named Coordinate-based Dual-constrained Autoregressive Motion Generation (CDAMD). We adopt joint coordinates as the motion representation and train two coordinate-based motion encoders: a Deterministic Autoencoder (AE) and a Residual Vector Quantized Variational Autoencoder (RVQ-VAE)~\cite{guo2024momask,DBLP:conf/cvpr/LeeKKCH22}. We propose Dual-Constrained Autoregressive Generation, where the Dual-Constrained Causal Mask is explicitly constrained along two orthogonal axes: temporal causality and conditional causality. Motion tokens derived from the RVQ-VAE serve as motion priors and are concatenated with textual encodings to compensate for the limitations of using only text conditions in capturing kinematic details. After the autoregressive decoding of motion latents, the generated tokens are refined by Diffusion Multi-Layer Perceptrons (Diffusion MLPs)~\cite{meng2025rethinking} to mitigate error accumulation and enhance motion fidelity.
The proposed CDAMD supports flexible and user-friendly motion editing via coordinate-based input. As shown in Fig.~\ref{fig:intro}(b), it achieves a low Fréchet Inception Distance (FID) score with real motions and a high text-aligned R-Precision, benefiting the Dual-Constrained Causal Mask. 

Our main contributions are as follows:
\begin{itemize}
	\item We propose a flexible, high-fidelity, and semantically faithful text-to-motion framework, which addresses limitations of both diffusion and autoregressive methods.
	\item We introduce the Dual-Constrained Causal Mask, which enforces both temporal and conditional causality under hybrid conditions of motion tokens and textual semantics.
	\item We establish coordinate-based motion synthesis benchmarks for both text-to-motion generation and editing.
\end{itemize}

\section{Related Works}

\subsection{Diffusion-Based Motion Generation} 
Diffusion models have recently become a leading framework for human motion generation~\cite{weng2026realign,taneasytune}. MDM~\cite{tevet2023human} combines diffusion with Transformer~\cite{vaswani2017attention} backbones and classifier-free guidance to produce high-quality 3D motions. Fg-T2M~\cite{wang2023fg} incorporates a linguistics-structure assisted module to better align textual descriptions with fine-grained body movements. MLD~\cite{chen2023executing} performs diffusion in the latent space rather than the high-dimensional motion space, thereby improving generation efficiency. ReMoDiffuse~\cite{zhang2023remodiffuse} enhances diffusion models with a retrieval module that integrates existing semantically similar motions into the denoising process. MoFusion~\cite{dabral2023mofusion} provides a general framework which supports multiple conditioning modalities, including text and music. sMDM~\cite{Bae_2025_ICCV} streamlines motion diffusion by attending to sparsely selected keyframes and interpolating intermediate frames. These diffusion-based motion models suffer from error accumulation during the noise prediction. 

\subsection{Autoregressive Motion Models} 
Apart from diffusion-based motion models, autoregressive motion modeling has also emerged as a popular direction for human motion generation. T2M-GPT~\cite{zhang2023generating} presents a two-stage autoregressive framework, consisting of VQ-VAE–based motion tokenization followed by GPT-like autoregressive generation. AttT2M~\cite{zhong2023attt2m} enhances autoregressive motion generation through a global-local motion-text attention mechanism. MotionGPT~\cite{jiang2023motiongpt} converts 3D motion sequences into discrete motion tokens and designs a unified transformer model to perform language modeling on both motion and text. By leveraging pre-trained large language models (LLMs), MotionGPT~\cite{zhang2024motiongpt} further interprets complex textual instructions and generates high-fidelity human motions. AMD~\cite{han2024amd} integrates autoregressive modeling with diffusion-based generation to combine the strengths of both paradigms. BAMM~\cite{pinyoanuntapong2024bamm} introduces a bidirectional autoregressive framework that combines generative masked modeling with autoregressive prediction. MotionStreamer~\cite{Xiao_2025_ICCV} combines a diffusion-based autoregressive model with a causal latent space to enable continuous, text-adaptive motion prediction. Although autoregressive motion models effectively avoid the issue of error amplification, they tend to suffer from mode collapse by repeating high-frequency tokens, as discretization results in the loss of fine-grained motion details. DisCoRD~\cite{cho2025discord} improves the naturalness of discrete motion generation by replacing the conventional token decoder with a conditional rectified-flow decoder in the continuous motion space. Its core idea is to treat discrete tokens generated by a pretrained token predictor as frame-wise conditioning signals for continuous motion decoding, thereby reducing frame-wise noise and under-reconstruction artifacts. In contrast, we incorporate motion priors directly into the autoregressive latent generation stage and introduce Dual-Constrained Causal Attention to regulate the visibility of text, motion priors, and generative positions. Diffusion Multi-Layer Perceptrons in our framework serves as a lightweight refinement module after autoregressive latent prediction, rather than the primary continuous generator. Therefore, while both methods combine discrete priors with continuous refinement, DisCoRD focuses on decoder-side naturalness enhancement, whereas our method targets condition-aware autoregressive generation and causal control.

\subsection{Human Motion Editing} 
Text-to-motion editing, also known as controllable human motion generation, is an active research focus. GMD~\cite{karunratanakul2023guided} introduces explicit spatial and temporal constraints such as keyframes, trajectories, or obstacles into motion diffusion. Hierarchical semantic graphs are employed to provide fine-grained control over motion generation~\cite{jin2023act}. DNO~\cite{karunratanakul2024optimizing} optimizes the latent noise of a pre-trained text-to-motion model to align with desired motion criteria. OmniControl~\cite{xieomnicontrol} enables control over any joint at any time, facilitating detailed and dynamic motion generation. MotionLCM~\cite{dai2024motionlcm} adopts one-step or few-step inference and incorporates a motion ControlNet within the latent space, allowing for explicit control signals. A multi-task paradigm is trained to jointly learn motion editing and motion similarity prediction~\cite{li2025simmotionedit}. DART~\cite{zhaodartcontrol} enables spatial control over generated motions by learning a compact motion primitive space within a diffusion-based autoregressive framework. MotionReFit~\cite{jiang2025dynamic} enhances text-driven motion editing by pairing MotionCutMix augmentation with an autoregressive diffusion model for more robust spatial–temporal edits. However, existing works use a mixed motion representation of local-relative joint positions and kinematic-aware 6D rotations, which does not align well with the conditions provided by human users.

\section{Motivation}

\subsection{Autoregressive Motion Diffusion}
Autoregressive motion diffusion models~\cite{han2024amd,zhaodartcontrol,meng2025rethinking} provide a promising paradigm for human motion modeling, addressing the limitations of both diffusion-based and autoregressive approaches. A representative model is MARDM~\cite{meng2025rethinking}, which is described below.

This model begins by reconstructing motion representations to improve adaptability. Instead of using redundant mixed motion representation encompassing 6D rotations and binary foot-contact indicators, it first retains only the essential 3D continuous feature groups, including root angular velocity, root linear velocities in the XZ-plane, root height, and local joint 3D positions. 
These features are then projected into a compact latent space through a deterministic 1D ResNet-based AutoEncoder (AE), where the decoder uses nearest-neighbor upsampling to reconstruct the motion features. 

The forward diffusion process generates noisy motion latents by progressively adding Gaussian noise $\epsilon$ to the clean motion $x_0$, following the formulation:
\begin{equation}
	x_t = \alpha_t x_0 + \sigma_t \epsilon,
	\label{eq:linear_interpolation}
\end{equation}
where $\alpha_t = 1 - t, \sigma_t = t$, and $t$ is the continuous timestep. 

The autoregressive diffusion architecture consists of a masked autoregressive transformer and diffusion MLPs. The transformer processes unmasked latent tokens, which are defined as $u$, using bidirectional attention to extract contextual information and produces conditional signals $z$ for the diffusion branch. During training, the cosine masking schedule~\cite{guo2024momask} randomly masks parts of the latent sequence, encouraging robust generation. 
Instead of directly predicting clean motion, the training objective combines coordinate prediction and velocity prediction losses. For noise prediction of coordinates, the model minimizes:
\begin{equation}
	\mathcal{L}_{C} = \mathbb{E}_{\epsilon, t}  \left\| \epsilon - \epsilon_{\theta} \bigl( \mathbf{x'}_t^{i} \big| t, g(u) \bigr) \right\|^2, 
	\label{eq:DDPM_loss}
\end{equation}
where $x'^i$ is motion latents produced by diffusion branch, $g(\cdot)$ is the autoregressive transformer, $\epsilon_{\theta}$ is the predicted noise conditioned on the transformer output $g(u)$, $\theta$ is the parameters of the diffusion MLPs. Similarly, for velocity-based loss, this model minimizes:
\begin{equation}
	\mathcal{L}_{V} = \int_{0}^{T} \mathbb{E}_{\mathbf{v}, t} \left\| \mathbf{v}_{\theta} \bigl( \mathbf{x'}_t^{i} \big| t, g(u) \bigr) - \dot{\alpha}_t \mathbf{x'}_0^{i} - \dot{\sigma}_t \epsilon \right\|^2  dt,
	\label{eq:v_loss}
\end{equation}
where $\mathbf{v}_{\theta}$ is the predicted velocity, $\alpha_t$ and $\sigma_t$ are continous time coefficients, $\dot{\alpha}_t = \frac{\mathrm{d}\alpha_t}{\mathrm{d}t}$, $\dot{\sigma}_t = \frac{\mathrm{d}\sigma_t}{\mathrm{d}t}$.

During sampling, masked vectors are progressively filled to reconstruct the entire latent sequence. The diffusion MLPs employ distinct sampling strategies for coordinates and velocities. For joint coordinates, the sampling process can be denoted as:
\begin{equation}
	\mathbf{x}_{t-1}^i = \frac{1}{\sqrt{\alpha_t}} \left( \mathbf{x}_t^i - \frac{\sqrt{1 - \alpha_t}}{\sqrt{1 - \bar{\alpha}_t}} \epsilon_\theta(\mathbf{x}_t^i \mid t, z^i) \right) + \sigma_t \epsilon_t,
	\label{eq:DDPM_sampling}
\end{equation}
where $\epsilon_t \sim \mathcal{N}(0, \mathbf{I})$ for noise prediction. Alternatively, for velocity prediction, ODE is employed to sample with step size $\Delta t$, given by:
\begin{equation}
	\mathbf{x}_{t-1}^i = \mathbf{x}_t^i + \Delta t \cdot \mathbf{v}_\theta \bigl( \mathbf{x}_t^i \big| t, z^i \bigr).
	\label{eq:v_sampling}
\end{equation}

\subsection{Beyond Temporal Causality}
Although autoregressive motion diffusion models provide a promising paradigm by combining the strengths of diffusion and autoregressive modeling, they suffer from several limitations. 
First, such methods enforce temporal causality through masking strategies, while overlooking the structured interaction between conditional tokens and generative tokens. This lack of conditional causality may result in unstable generation and suboptimal semantic alignment. 
Second, the conditioning textual descriptions are often insufficient to capture fine-grained motion dynamics, while masked latent tokens do not constitute a reliable motion prior.

These limitations motivate us to design a dual-constrained autoregressive motion generation, where both temporal and conditional causality are explicitly modeled. 
In addition, we introduce discrete motion tokens as structured motion priors to complement textual semantics, enabling more faithful and controllable motion generation.

\section{Method}
\label{method}
\begin{figure*}[htbp]    
	\centering    
	\includegraphics[height=7cm, width=\textwidth]{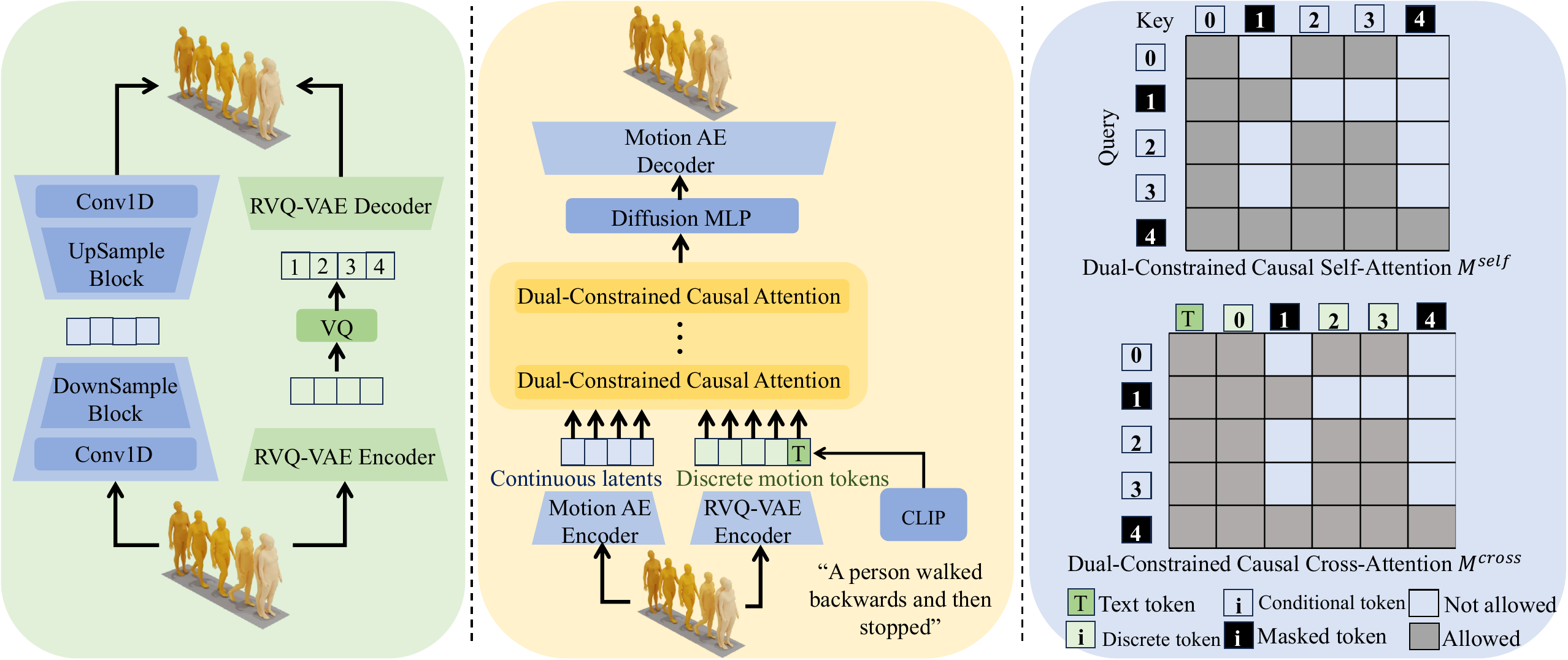}    
	\caption{Architecture illustration of CDAMD. (a) Hybrid Motion Encoders encodes the raw motion sequence into a compact fine-grained latent space. (b) CDAMD model learns to autoregressively predict next tokens conditioned on text embedding from CLIP and compressed motion tokens from RVQ-VAE. (c) Dual-Constrained Causal Attention (DCCA) enforces both temporal and conditional causality, ensuring that motion generation proceeds autoregressively and preserving semantic conditioning.}    
	\label{fig:overview}
\end{figure*}

We introduce a text-to-motion framework, named CDAMD, that achieves both high-fidelity, semantically consistent motion synthesis and inherent editability for fine-grained control. The proposed framework is conditioned on a textual prompt $c_{t}$, which is first encoded into an embedding $e_{t}$ using the CLIP, and the discrete motion tokens $H$. To unify motion generation and editing, we introduce a Dual-Constrained Causal Attention mechanism and augment the autoregressive model with a controllable binary mask. During training, the model is optimized via the following objective:
\begin{equation}    
	\mathcal{L}_{D} = \mathbb{E}_{z_0, \epsilon \sim \mathcal{N}(0, I), s, c_{t}, H, M}  \| \epsilon - \epsilon_\theta(z_s, s, e_{t}, H, M) \|_2^2, 
\end{equation}
where $M=\{m_t\}_{t=1}^l$ denotes a binary mask, with $m_t=1$ marking tokens to be generated and $m_t=0$ marking known conditioning tokens. During inference, the model starts from pure noise $z_S \sim \mathcal{N}(0, I)$ and iteratively denoises it, guided by the mask $M$ and the conditions of $e_t$ and $H$, to produce the final latent sequence $Z$. 
Architecture of the CDAMD is illustrated in Fig.~\ref{fig:overview}, and details are described below.

\subsection{Coordinate-Based Motion Encoders}
We employ joint coordinates as the motion representation, following the recent practices~\cite{meng2025rethinking,meng2025absolute}. To fully represent joint coordinates in the latent space, we design two coordinate-based motion encoders: Deterministic Autoencoder (AE) and Residual Vector Quantized Variational Autoencoder (VAE).

\subsubsection{Deterministic AE}
To model the generation process on a lower-dimensional and smoother manifold, we employ an autoencoder to learn a compact representation of motion. The AE consists of an encoder $\mathcal{E}_{ae}$ and a decoder $\mathcal{D}_{ae}$. The encoder maps the raw motion sequence $\mathcal{X}$ to a continuous latent sequence of length $l<L$, denoted as $Z = \{z_t\}_{t=1}^l$, where each $z_t \in \mathbb{R}^{d}$ is a $d$-dimensional latent vector. We choose a deterministic AE to eliminate latent stochasticity, thereby providing a stable representation for the diffusion process. The AE is trained by minimizing the reconstruction loss:
\begin{equation}
	\mathcal{L}_{AE} = \mathbb{E}_{\mathcal{X} \sim p_{data}(\mathcal{X})}  \| \mathcal{X} - \mathcal{D}_{ae}(\mathcal{E}_{ae}(\mathcal{X})) \|_1,
	\label{eq:ae_loss}
\end{equation}
where $p_{data}$ is the distribution of real-world motion data. This latent sequence $Z$ serves as the target representation for our diffusion-based generative model.

\subsubsection{Residual Vector Quantized VAE}
To further furnish a powerful discrete motion prior and provide symbolic conditioning, we additionally train a Residual Vector Quantized Variational Autoencoder (RVQ-VAE)~\cite{guo2024momask}. Unlike the deterministic AE, which produces continuous latents, RVQ-VAE discretizes the latent space into a finite set of codewords, effectively filtering redundancies and extracting the essential dynamics of motion. Formally, the encoder $\mathcal{E}_{vq}$ maps the input sequence $\mathcal{X}$ into latent embeddings $Z = \{z_t\}_{t=1}^l$. 
The decoder $\mathcal{D}_{vq}$ reconstructs the motion sequence as $\hat{\mathcal{X}} = \mathcal{D}_{vq}(\hat{Z})$. 
Each latent $z_t$ is then quantized via a multi-level residual codebook $\mathcal{C} = \{\mathcal{C}^{1}, \dots, \mathcal{C}^{R}\}$, where $R$ is the number of quantization levels. At the $r$-th level, the nearest codeword is selected as:
\begin{equation}
	e^{r} = \arg\min_{c \in \mathcal{C}^{r}} \| z^{r-1} - c \|_2^2, 
	\label{eq:vq_nearest}
\end{equation}
where $z^{0} = z_t$ and $z^{r} = z^{r-1} - e^{r}$.
The quantized embedding is reconstructed by summing over residuals:
\begin{equation}
	\hat{z}_t = \sum_{r=1}^{R} e^{r}.
\end{equation}

The training process optimizes the quantization using the standard vector quantization loss with a commitment term:
\begin{equation}
	\mathcal{L}_{VQ} = \| \text{sg}(\mathbf{z}) - \mathbf{e} \|_2^2 + \beta \| \mathbf{z} - \text{sg}(\mathbf{e}) \|_2^2,
	\label{eq:vq_loss}
\end{equation}
where $\text{sg}(\cdot)$ denotes the stop-gradient operator, $e$ is the selected codeword, and $\beta$ is a hyperparameter controlling the strength of the commitment loss. 

\subsection{Dual-Constrained Causal Attention}
To elaborate Dual-Constrained Causal Attention, we first introduce Conditional Motion Priors and Hybrid Conditions with Motion Tokens.

\subsubsection{Conditional Motion Priors}
A key advantage of RVQ-VAE lies not only in compression but also in the provision of discrete motion tokens $H = \{h_t\}_{t=1}^l$, which serve as motion priors. These tokens capture domain-specific kinematic structures that are difficult to infer from textual descriptions alone. We concatenate motion tokens $H$ with text embeddings, forming hybrid conditioning sequences that are fed into the transformer. This design alleviates the limited guidance capacity of text-only inputs in motion generation, enabling the model to better align semantics with physically plausible motion dynamics. Thus, discrete tokens act as symbolic anchors that guide the generative process, complementing the high-level semantics provided by the language.

\subsubsection{Hybrid Conditions with Motion Tokens} 
To avoid overreliance on motion priors, the discrete tokens are stochastically perturbed during training. With a probability of 70\%, the entire motion token stream is discarded, forcing the model to rely more strongly on textual guidance. For the remaining valid latents, a cosine-scheduled masking strategy is applied, where the masked tokens are replaced by Gaussian noise (10\%) or a learnable embedding [MASK] (88\%). This hybrid perturbation balances the reliance on textual semantics and motion continuity, preventing overfitting and improving generalization.

\subsubsection{Dual-Constrained Causal Attention} 
The transformer backbone is constructed from Dual-Constrained Causal Attention (DCCA) blocks, where attention is explicitly constrained along two orthogonal axes: temporal causality and conditional causality. The transformer backbone consists of two types of DCCA blocks: Dual-Constrained Causal Self-Attention (DCCA-S) and Dual-Constrained Causal Cross-Attention (DCCA-C).

In the DCCA-S, latent-to-latent interactions are governed by a dual mask. Let $X \in \mathbb{R}^{L \times d}$ denote the latent sequence, with temporal and conditional masks $M^{\text{temp}}$ and $M^{\text{cond}}$. The self-attention mask is defined as:
\begin{equation}
	M^{\text{self}} = M^{\text{temp}} \cap M^{\text{cond}},
\end{equation}
where \( M^{\text{temp}}(i, j) = 1 \) if \( j \leq i \), disallowing the
$i$-th token from accessing future tokens, and \( M^{\text{cond}}(i, j) = 1 \) if \( j \) corresponds to a condition position that is globally visible or \( j \leq i \) otherwise. The corresponding self-attention operation is as follows:
\begin{equation}
	\text{DCCA-S}(X) = \text{softmax}\left( \frac{Q K^\top}{\sqrt{d}} + \log M^{\text{self}} \right) V,
\end{equation}
where \( Q, K, V \) are the query, key, and value projections. This ensures that generative positions only attend to the past, while condition positions are globally accessible but isolated from generative ones.

In the DCCA-C, latent-to-condition attention operates over text tokens \( T \in \mathbb{R}^{N \times d} \) and motion tokens \( Z \in \mathbb{R}^{M \times d} \). Text tokens are globally visible, while motion tokens inherit the same dual-constrained rules as latents. The cross-attention mask is defined as:
\begin{equation}
	M^{\text{cross}}(i, j) =
	\begin{cases}
		1, & j \in T,  \\
		M^{\text{self}}(i, j), & j \in Z.
	\end{cases}
\end{equation}
The resulting cross-attention operation is formulated as:
\begin{equation}
	\text{DCCA-C}(X, [T, Z]) = \text{softmax}\left( \frac{Q K^\top}{\sqrt{d}} + \log M^{\text{cross}} \right) V.
\end{equation}
By jointly enforcing causality along the temporal axis and conditional separation across different modalities, DCCA guarantees strict autoregressive consistency while preserving semantic fidelity.

\subsection{Autoregressive Motion Generation}
The transformer backbone, built with DCCA blocks, governs this generation process. Latent positions strictly obey temporal causality, ensuring that future slots remain inaccessible, while conditional slots are globally visible but separated from generative slots. Cross-attention injects cross-modal information: text tokens are always accessible, whereas motion tokens follow the same dual causality rules as latents. This dual-constrained mechanism ensures that generation proceeds in temporal order, consistently guided by both semantic and motion priors.

After the autoregressive decoding of latents, the sequence is refined by a DiffMLPs sampler. Depending on configuration, Diffusion Multi-Layer Perceptrons (Diffusion MLPs) implements either a DDPM~\cite{ho2020denoising}-style denoising sampler or a SiT~\cite{ma2024sit}-style transport sampler. Given the autoregressive outputs \( \mathbf{z} \), the diffusion stage reconstructs motion latents through iterative denoising or ODE-based refinement in SiT. This step mitigates error accumulation and aligns generated motions with the training distribution, enhancing realism and fidelity. Finally, the refined motion latents are decoded into continuous 3D motion sequences using the pretrained AE decoder.

\subsubsection{Training} 
The Conditional Motion Prior (CMP), instantiated by the discrete motion tokens derived from the RVQ-VAE, is used during training as a training-time motion prior. Specifically, for each training motion sequence, we first encode the raw coordinates into continuous latents with the pretrained AE encoder and, in parallel, extract discrete motion tokens with the pretrained RVQ-VAE encoder. The transformer is then trained to predict masked continuous latents under hybrid conditions consisting of the text embedding and the discrete motion priors. The motion tokens act as teacher priors that provide additional kinematic structure beyond language, helping the model learn a more stable and motion-aware latent denoising process.

To prevent the model from over-relying on these training-time priors, we apply a strong stochastic dropout strategy to the motion-token stream. With a probability of 70\%, the entire motion-token condition is removed during training, forcing the model to solve the task from text alone. For the remaining cases, the latent sequence is perturbed with the same masking strategy described above, where masked positions are replaced by Gaussian noise or a learnable [MASK] embedding. This design explicitly reduces the discrepancy between training and inference, since the model is repeatedly exposed to text-only supervision during optimization rather than learning under a strictly stronger conditioning signal.

\subsubsection{Inference} 
During inference, we use only the textual description as the condition. The input is initialized as a fully masked latent sequence together with the CLIP text embedding, and the model progressively fills the latent positions in an autoregressive manner under the proposed dual-constrained causal masks. The predicted latent sequence is then refined by Diffusion MLPs and finally decoded by the pretrained AE decoder to obtain the output motion.

\section{Experiments}
\label{experiments}

\subsection{Datasets and Implementation Details}

\subsubsection{Datasets}
We conduct experiments on two widely used text-to-motion generation benchmarks: HumanML3D~\cite{Guo_2022_CVPR}  and KIT-ML~\cite{doi:10.1089/big.2016.0028}. The KIT-ML dataset consists of 3,911 motion sequences collected from the KIT and CMU~\cite{1571417125676818048} motion databases, each paired with one to four textual descriptions, resulting in a total of 6,278 annotations. All motion sequences are standardized to 12.5 FPS. The HumanML3D dataset contains 14,616 motion sequences derived from the AMASS~\cite{AMASS:2019} and HumanAct12~\cite{guo2020action2motion} datasets, each annotated with three textual descriptions, yielding 44,970 annotations in total. Motion sequences in HumanML3D are normalized to 20 FPS and clipped to a maximum length of 10 seconds. Following the standard setting, we apply motion mirroring for data augmentation and split them into training, validation, and test sets with respective proportions of 80\%, 5\%, and 15\%.

\subsubsection{Implementation Details}
For motion representation learning, we adopt an AutoEncoder (AE) and a residual vector quantized VAE (RVQ-VAE). 
The AE is a 3-layer convolutional encoder-decoder with hidden width $512$, depth $3$, dilation growth rate $3$, and a total temporal downsampling factor of $4$. 
For the VQ model, we employ a residual quantizer with $4$ quantization layers, a codebook size of $512$, and embedding dimension $512$, following the multi-stage quantization scheme. 
The AE produces continuous latent sequences, while the RVQ-VAE discretizes motion into a sequence of motion tokens serving as conditions.

The generation branch is based on a Masked Transformer. 
We utilize a transformer encoder with $8$ layers, hidden dimension $1024$, $6$ attention heads, feed-forward size $4096$, and dropout $0.1$.
We integrate a strict dual-causal cross-attention mechanism to enforce temporal causality: generation positions obey autoregressive masking, while condition positions (text and motion tokens) follow stricter visibility rules.
Additionally, we incorporate Diffusion MLPs (DiffMLPs) with a SiT-XL backbone, consisting of residual MLP blocks mapping the Transformer hidden dimension to the autoencoder latent space.

The learning rate is initialized at $2\times 10^{-4}$ with a linear warm-up of $2000$ iterations, followed by a cosine annealing schedule with a minimum learning rate of $1\times 10^{-6}$. 
For the HumanML3D, we train with batch size $64$, and for the KIT-ML with batch size $16$, both with a maximum motion length of $196$ frames. 
The AE and RVQ-VAE are pre-trained separately for $50$ epochs before being fixed during the training of the Transformer. 
During Transformer training, we set the maximum training epochs to $500$. 

\subsection{Experimental Setup}
\subsubsection{Coordinate-Based Evaluation Metrics} 
Since most existing motion generation works use joint rotation angles rather than joint coordinates, there is a lack of coordinate-based text-to-motion benchmarks. We train evaluators using absolute joint coordinates: one following the architecture proposed in T2M~\cite{Guo_2022_CVPR} and another based on CLIP~\cite{tevet2022motionclip}, and use them to evaluate different methods.

For the T2M evaluator, we adopt standard metrics: (1) R-Precision (evaluated at Top-1, Top-2, and Top-3) and Matching, which assess semantic alignment between motion and text embeddings; (2) Fréchet Inception Distance (FID), which measures distributional similarity between generated and real motions; and (3) MultiModality, which quantifies intra-text diversity across multiple generated samples conditioned on the same description. For the CLIP-based evaluator, we report the CLIP-Score~\cite{hessel2021clipscore}, defined as the similarity between motion and text embeddings, to capture cross-modal compatibility.

\subsection{Evaluation of Motion Synthesis}
\subsubsection{Results of Text-to-Motion Generation} 
We evaluate our model on the HumanML3D dataset and compare it with state-of-the-art methods. The quantitative results are presented in Table \ref{tab:quantitative evaluation on HumanML3D}. Under the coordinate-based setting, our model achieves the best overall fidelity-semantic trade-off. The best CDAMD variants reduce FID to 0.046, outperforming all listed baselines, including MoMask at 0.047, BAMM at 0.060, MMM at 0.093, MotionLCM V2 at 0.152, and MARDM-SiT-XL at 0.156. At the same time, CDAMD with the first VQ achieves the best Top-1 R-Precision of 0.522 and the best Matching score of 2.966, while the two- and all-VQ variants maintain similar retrieval performance with substantially better FID. In contrast, ACMDM-S-PS22 obtains the highest Diversity of 10.043, but its FID remains notably worse at 0.107. These results suggest that the CDAMD improves motion fidelity without a severe loss of diversity.


We also compare results using different numbers of motion quantizers in RVQ-VAE. Moving from the first VQ to two VQ levels yields the largest improvement in fidelity. Specifically, FID drops from 0.090 to 0.052, and under the coordinate-based setting, it drops from 0.082 to 0.046. However, adding all VQ levels does not bring further gains in FID, which stays at 0.051 in the standard setting and 0.046 in the coordinate-based setting, while Top-1 R-Precision and Matching degrade slightly relative to the shallower variants. Diversity also peaks at two VQ levels in the standard setting, reaching 9.911, before decreasing to 9.835 with all VQ levels. This pattern suggests that moderate prior strength is most beneficial, whereas deeper quantization may introduce redundant information that no longer improves conditional generation.

Text-to-motion generation results on the KIT-ML dataset are provided in Table~\ref{tab:quantitative evaluation on KIT}. Similar to the results on the HumanML3D dataset, we establish a coordinate-based text-to-motion generation benchmark on the KIT-ML dataset. Our approach consistently outperforms most existing methods across different evaluation metrics.

\begin{table*}[tbp]
	\centering
	\caption{Evaluation of text-to-motion generation on the HumanML3D datasets. More than 10 baselines are re-implemented using joint coordinates to reproduce their results on this dataset. The evaluation is repeated 20 times, and the average is reported with a 95\% confidence interval. \textbf{Bold} indicates the best result, and \underline{underscore} denotes the second best. $^*$ refers the result of our re-implementation. VQ denotes the motion quantizers in the Residual Vector Quantized Variational Autoencoder (RVQ-VAE).}
	\label{tab:quantitative evaluation on HumanML3D}
	\resizebox{\textwidth}{!}{   
		\begin{tabular}{l|c|ccc|c|c|c}
			\hline
			\multirow{2}{*}{Methods} & \multirow{2}{*}{FID$\boldsymbol{\downarrow}$} & \multicolumn{3}{c|}{R-Precision$\boldsymbol{\uparrow}$} & \multirow{2}{*}{Matching$\boldsymbol{\downarrow}$} & \multirow{2}{*}{MModality$\boldsymbol{\uparrow}$} & \multirow{2}{*}{Diversity$\boldsymbol{\uparrow}$} \\
			\cline{3-5}
			& & Top 1 & Top 2 & Top 3 & & \\
			\hline
			MDM-50Step$^*$~\cite{tevet2023human} & $0.395^{\pm .065}$ & $0.435^{\pm .013}$ & $0.627^{\pm .0014}$ & $0.737^{\pm .011}$ & $3.443^{\pm .060}$ & $2.182
			^{\pm .055}$ & $9.812^{\pm .148}$ \\
			MotionDiffuse$^*$~\cite{zhang2022motiondiffuse} & $1.334^{\pm .035}$ & $0.435^{\pm .008}$ & $0.618^{\pm .006}$ & $0.725^{\pm .005}$ & $3.542^{\pm .015}$ & $1.833^{\pm .078}$ & $9.634^{\pm .048}$ \\
			ReMoDiffuse$^*$~\cite{zhang2023remodiffuse} & $0.207^{\pm .006}$ & $0.467^{\pm .001}$ & $0.654^{\pm .004}$ & $0.748^{\pm .002}$ & $3.289^{\pm .022}$ & $\textbf{2.560}^{\pm .289}$ & $9.483^{\pm .245}$ \\
			MotionLCM V2$^*$~\cite{dai2024motionlcm} & $0.152^{\pm .007}$ & $0.511^{\pm .007}$ & $\underline{0.707}^{\pm .003}$ & $\textbf{0.802}^{\pm .002}$ & $3.005^{\pm .009}$ & $1.993^{\pm .085}$ & $9.703^{\pm .097}$ \\
			MMM$^*$~\cite{pinyoanuntapong2024mmm} & $0.093^{\pm .004}$ & $0.495^{\pm .008}$ & $0.687^{\pm .000}$ & $0.782^{\pm .003}$ & $3.165^{\pm .037}$ & $1.303^{\pm .085}$ & $9.786^{\pm .352}$ \\
			MoMask$^*$~\cite{guo2024momask} & $\underline{0.047}^{\pm .003}$ & $0.493^{\pm .002}$ & $0.686^{\pm .002}$ & $0.784^{\pm .002}$ & $3.124^{\pm .008}$ & $1.356^{\pm .042}$ & $\underline{9.838}^{\pm .074}$ \\
			BAMM$^*$~\cite{pinyoanuntapong2024bamm} & $0.060^{\pm .007}$ & $0.495^{\pm .003}$ & $0.690^{\pm .002}$ & $0.796^{\pm .002}$ & $3.148^{\pm .008}$ & $1.950^{\pm .067}$ & $9.343^{\pm .068}$ \\
			MARDM-SiT-XL$^*$~\cite{meng2025rethinking} & $0.156^{\pm .007}$ & $0.486^{\pm .003}$ & $0.680^{\pm .003}$ & $0.780^{\pm .002}$ & $3.136^{\pm .010}$ & $\underline{2.353}^{\pm .101}$ & $9.777^{\pm .082}$ \\
			ACMDM-S-PS22$^*$~\cite{meng2025absolute} & $0.107^{\pm .014}$ & $0.509^{\pm .002}$ & $0.699^{\pm .003}$ & $0.793^{\pm .003}$ & $3.036^{\pm .010}$ & $2.069^{\pm .061}$ & $\textbf{10.043}^{\pm .097}$ \\
			\textbf{CDAMD (w/ first VQ)} & $0.082^{\pm .003}$ & $\textbf{0.522}^{\pm .003}$ & $\textbf{0.708}^{\pm .002}$ & $\underline{0.800}^{\pm .002}$ & $\textbf{2.966}^{\pm .010}$ & $1.678^{\pm .058}$ & $9.680^{\pm .082}$ \\
			\textbf{CDAMD (w/ two VQ)} & $\textbf{0.046}^{\pm .003}$ & $\underline{0.521}^{\pm .002}$ & $0.705^{\pm .003}$ & $0.798^{\pm .002}$ & $\underline{2.975}^{\pm .008}$ & $1.704^{\pm .059}$ & $9.822^{\pm .081}$ \\
			\textbf{CDAMD (w/ all VQ)} & $\textbf{0.046}^{\pm .002}$ & $0.518^{\pm .003}$ & $0.704^{\pm .003}$ & $0.794^{\pm .002}$ & $2.998^{\pm .010}$ & $1.674^{\pm .058}$ & $9.793^{\pm .087}$ \\
			\hline
		\end{tabular}
	}
\end{table*}

\begin{table*}[tbp]
	\centering
	\caption{Evaluation of coordinate-based text-to-motion generation on the KIT dataset. Baselines are re-implemented using joint coordinates to reproduce their results on this dataset. \textbf{Bold} indicates the best result, and \underline{underscore} denotes the second best.
	}
	\label{tab:quantitative evaluation on KIT}
	\resizebox{0.9\textwidth}{!}{  
		\begin{tabular}{l|c|ccc|c|c|c}
			\hline
			\multirow{2}{*}{Methods} & \multirow{2}{*}{FID$\boldsymbol{\downarrow}$} & \multicolumn{3}{c|}{R-Precision$\boldsymbol{\uparrow}$} & \multirow{2}{*}{Matching$\boldsymbol{\downarrow}$} & \multirow{2}{*}{MModality$\boldsymbol{\uparrow}$} & \multirow{2}{*}{Diversity$\boldsymbol{\uparrow}$} \\
			\cline{3-5}
			& & Top 1 & Top 2 & Top 3 & & \\
			\hline
			MDM~\cite{tevet2023human} & $0.782^{\pm .066}$ & $0.390^{\pm .007}$ & $0.609^{\pm .010}$ & $0.733^{\pm .012}$ & $3.956^{\pm .110}$ & $\textbf{2.402}^{\pm .257}$ & $11.019^{\pm .139}$ \\
			MotionDiffuse~\cite{zhang2022motiondiffuse} & $2.793^{\pm .192}$ & $0.357^{\pm .010}$ & $0.553^{\pm .009}$ & $0.667^{\pm .012}$ & $3.957^{\pm .062}$ & $1.538^{\pm .062}$ & $11.316^{\pm .153}$ \\
			ReMoDiffuse~\cite{zhang2023remodiffuse} & $\textbf{0.257}^{\pm .021}$ & $\underline{0.412}^{\pm .006}$ & $\underline{0.630}^{\pm .012}$ & $\underline{0.754}^{\pm .009}$ & $\textbf{3.097}^{\pm .043}$ & $1.814^{\pm .103}$ & $\underline{11.451}^{\pm .183}$ \\
			MoMask~\cite{guo2024momask} & $0.523^{\pm .022}$ & $0.392^{\pm .006}$ & $0.604^{\pm .008}$ & $0.732^{\pm .006}$ & $3.383^{\pm .030}$ & $\underline{1.892}^{\pm .085}$ & $11.143^{\pm .072}$ \\
			\hline
			\textbf{CDAMD (ours)} & $\underline{0.270}^{\pm .020}$ & $\textbf{0.416}^{\pm .006}$ & $\textbf{0.632}^{\pm .005}$ & $\textbf{0.759}^{\pm .004}$ & $\underline{3.160}^{\pm .027}$ & $1.211^{\pm .075}$ & $\textbf{11.902}^{\pm .154}$ \\
			\hline
		\end{tabular}
	}
\end{table*}

\begin{table}[tbp]
	\centering
	\caption{Results of temporal motion editing on the HumanML3D. We use joint coordinates as motion representation, and re-implement results of BAMM~\cite{pinyoanuntapong2024bamm} and MARDM~\cite{meng2025rethinking}.}
	\label{tab:edit evaluation on HumanML3D}
	\resizebox{0.5\textwidth}{!}{   
		\begin{tabular}{c|l|ccc|c|c|c}
			\hline
			\multirow{2}{*}{Tasks} & \multirow{2}{*}{Methods} & \multicolumn{3}{c|}{R-Precision$\boldsymbol{\uparrow}$} & \multirow{2}{*}{FID$\boldsymbol{\uparrow}$} & \multirow{2}{*}{Matching$\boldsymbol{\downarrow}$} & \multirow{2}{*}{CLIP-score$\boldsymbol{\uparrow}$} \\
			\cline{3-5}
			& & Top 1 & Top 2 & Top 3 & & \\
			\hline
			\multirow{3}{*}{\makecell{Temporal \\Inpainting}} & BAMM~\cite{pinyoanuntapong2024bamm} & 0.387 & 0.554 & 0.649 & 0.385 & 4.046 & 0.574 \\
			& MARDM~\cite{meng2025rethinking} & \textbf{0.503} & \textbf{0.702} & \textbf{0.795} & 0.120 & 3.051 & \textbf{0.671} \\
			& \textbf{CDAMD (ours)} & 0.500 & 0.691 & 0.790 & \textbf{0.103} & \textbf{3.049} & 0.665 \\
			\hline
			\multirow{3}{*}{\makecell{Temporal \\Outpainting}} & BAMM~\cite{pinyoanuntapong2024bamm} & 0.433 & 0.605 & 0.707 & 0.206 & 3.615 & 0.613 \\
			& MARDM~\cite{meng2025rethinking} & 0.512 & 0.705 & 0.797 & 0.114 & 3.065 & 0.671 \\
			& \textbf{CDAMD (ours)} & \textbf{0.523} & \textbf{0.717} & \textbf{0.809} & \textbf{0.104} & \textbf{2.916} & \textbf{0.676} \\
			\hline
			\multirow{3}{*}{\makecell{Temporal \\Prefix}} & BAMM~\cite{pinyoanuntapong2024bamm} & 0.352 & 0.526 & 0.632 & 0.578 & 4.178 & 0.565 \\
			& MARDM~\cite{meng2025rethinking} & \textbf{0.515} & \textbf{0.709} & \textbf{0.800} & \textbf{0.120} & \textbf{3.039} & \textbf{0.673} \\
			& \textbf{CDAMD (ours)} & 0.490 & 0.681 & 0.782 & 0.163 & 3.086 & 0.663 \\
			\hline
			\multirow{3}{*}{\makecell{Temporal \\Suffix}} & BAMM~\cite{pinyoanuntapong2024bamm} & 0.435 & 0.608 & 0.708 & 0.201 & 3.504 & 0.625 \\
			& MARDM~\cite{meng2025rethinking} & 0.501 & 0.685 & 0.780 & 0.138 & 3.113 & 0.668 \\
			& \textbf{CDAMD (ours)} & \textbf{0.541} & \textbf{0.730} & \textbf{0.820} & \textbf{0.090} & \textbf{2.848} & \textbf{0.679} \\
			\hline
		\end{tabular}
	}
\end{table}

\begin{table}[tbp]
	\centering
	\caption{Comparison between different masking strategies in the transformer backbone. We compare the proposed Dual-Constrained Causal Mask (DCCM) with the Bidirectional Causal Mask (BCM)~\cite{pinyoanuntapong2024bamm} and the Causal Mask (CM) on the HumanML3D dataset.}
	\label{tab:mask_first_quant}
	\resizebox{0.5\textwidth}{!}{   
		\begin{tabular}{l|c|ccc|c|c|c}
			\hline
			\multirow{2}{*}{Methods} & \multirow{2}{*}{FID$\boldsymbol{\downarrow}$} & \multicolumn{3}{c|}{R-Precision$\boldsymbol{\uparrow}$} & \multirow{2}{*}{Matching$\boldsymbol{\downarrow}$} & \multirow{2}{*}{Diversity$\boldsymbol{\uparrow}$}  & \multirow{2}{*}{CLIP-score$\boldsymbol{\uparrow}$}\\
			\cline{3-5}
			& & Top 1 & Top 2 & Top 3 & & \\
			\hline
			DCCM & \textbf{0.082} & \textbf{0.522} & 0.708 & 0.800 & 2.966 & 9.680 & 0.679 \\
			BCM & 0.154 & 0.520 & \textbf{0.710} & \textbf{0.805} & \textbf{2.925} & \textbf{9.770} & \textbf{0.683}\\
			CM & 0.231 & 0.515 & 0.696 & 0.792 & 3.055 & 9.132 & 0.671 \\
			\hline
		\end{tabular}
	}
\end{table}

\subsubsection{Results of Temporal Motion Editing}
For temporal editing tasks, the conditioning signal consists of both the text description and the observed motion segments. Specifically, given a ground-truth motion sequence $X$, we first encode it with the pretrained AE encoder to obtain latent tokens $Z \in \mathbb{R}^{l \times d}$. Editing is then performed in the latent space rather than in the raw coordinate space. A task-specific binary mask $M \in \{0,1\}^{l}$ is constructed, where $M_t=1$ denotes positions to be edited and $M_t=0$ denotes observed positions that remain as conditions. We conduct experiments on four temporal editing tasks, namely temporal inpainting (motion in-betweening), temporal outpainting, prefix, and suffix. Inpainting is evaluated by generating the middle 50\% of a motion sequence given its first and last 25\%. Outpainting performs the reverse operation. For prefix generation, the model is conditioned on the first 50\% of the ground-truth motion to predict the remaining part, while suffix generation does the opposite. We re-implement results of BAMM~\cite{pinyoanuntapong2024bamm} and MARDM~\cite{meng2025rethinking} for coordinate-based motion editing, and results on the HumanML3D dataset are shown in Table~\ref{tab:edit evaluation on HumanML3D}. 
Our approach significantly outperforms BAMM on all motion editing tasks, and surpasses MARDM for most tasks such as temporal outpainting and temporal suffix.

\subsection{Ablation Studies and Analysis}

\subsubsection{Comparison of Masking Mechanisms} 
To further analyze the role of attention masking strategies, we compare the proposed Dual-Constrained Causal Mask with the commonly used Bidirectional Causal Mask. Quantitative results are reported in Table~\ref{tab:mask_first_quant}.
We observe that the proposed Dual-Constrained Causal Mask achieves a significantly lower FID, indicating superior fidelity. The Top-1 R-Precision is also slightly higher, showing that the generated motions are better matched with the text descriptions. On the other hand, the Bidirectional Causal Mask achieves marginally better scores on Top-2 and Top-3 R-Precision, Matching, Diversity, and CLIP-score, suggesting that allowing full visibility of conditions can lead to more diverse generations and a slight improvement in semantic alignment at coarser levels.
These results highlight a trade-off between the two masking strategies. While the Bidirectional Causal Mask facilitates higher diversity and broader semantic coverage, our Dual-Constrained Causal Mask enforces stricter temporal alignment between conditions and generated tokens, thereby yielding more faithful motion distributions.

\begin{table*}[tbp]
	\centering
	\caption{Ablation study of the design choices of CDAMD on the HumanML3D dataset. CMP and DCCA denote Conditioned Motion Prior and Dual-Constrained Causal Attention, respectively. $^{\dagger}$ denotes architectural components under the standard 263-dimensional mixed representation.} 
	\label{tab:ablation study on HumanML3D}
	\resizebox{0.8\textwidth}{!}{ 
		\begin{tabular}{l|c|ccc|c|c|c}
			\hline
			\multirow{2}{*}{Methods} & \multirow{2}{*}{FID$\boldsymbol{\downarrow}$} & \multicolumn{3}{c|}{R-Precision$\boldsymbol{\uparrow}$} & \multirow{2}{*}{Matching$\boldsymbol{\downarrow}$} & \multirow{2}{*}{Diversity$\boldsymbol{\uparrow}$}  & \multirow{2}{*}{CLIP-score$\boldsymbol{\uparrow}$}\\
			\cline{3-5}
			& & Top 1 & Top 2 & Top 3 & & \\
			\hline
			\textbf{CDAMD (ours)} & \textbf{0.082} & \textbf{0.522} & \textbf{0.708} & \textbf{0.800} & \textbf{2.966} & \textbf{9.680} & \textbf{0.679} \\
			w/o CMP & 0.136 & 0.520 & 0.704 & 0.796 & 3.015 & 9.495 & 0.676\\
			w/o DCCA & 0.231 & 0.515 & 0.696 & 0.792 & 3.055 & 9.132 & 0.671 \\
			\hline
			\textbf{CDAMD (ours)}$^{\dagger}$ & \textbf{0.090} & \textbf{0.523} & \textbf{0.712} & \textbf{0.808} & \textbf{2.942} & \textbf{9.728} & \textbf{0.684} \\
			w/o CMP$^{\dagger}$ & 0.122 & 0.516 & 0.699 & 0.791 & 3.109 & 9.686 & 0.681\\
			w/o DCCA$^{\dagger}$ & 0.159 & 0.510 & 0.687 & 0.776 & 3.256 & 9.384 & 0.672 \\
			\hline
		\end{tabular}
	}
\end{table*}

\begin{figure*}[tbp]            
	\centering            
	\includegraphics[width=0.95\textwidth]{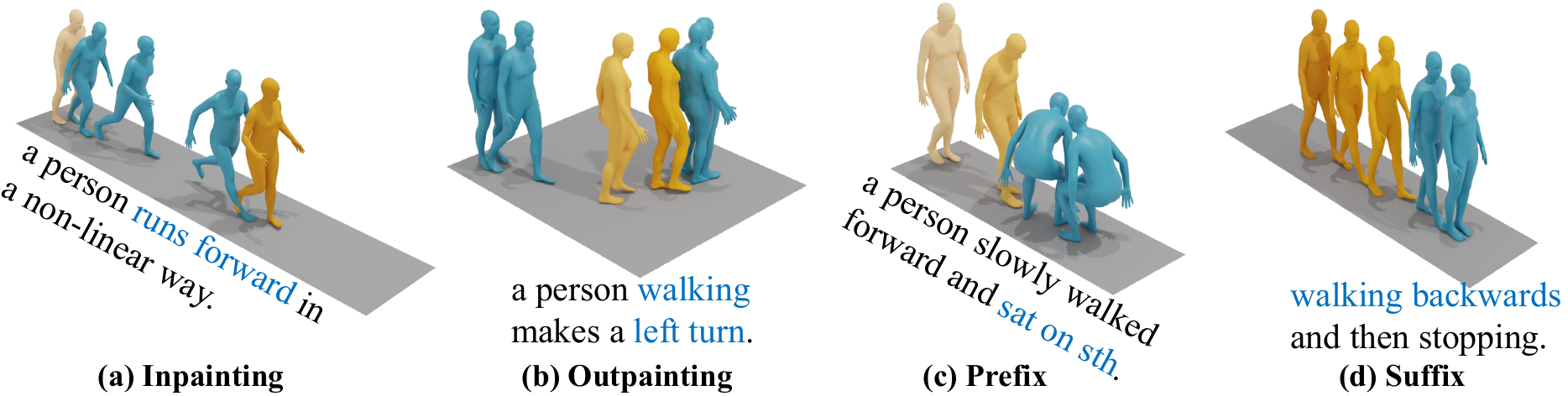}            
	\caption{Visualization of temporal editing tasks, inpainting, outpainting, prefix, and suffix where orange indicates conditioned motion and blue refers to generated parts.}       
	\label{fig:edit}
\end{figure*}

\begin{figure*}[tbp]    
	\centering    
	\includegraphics[height=9cm, width=\textwidth]{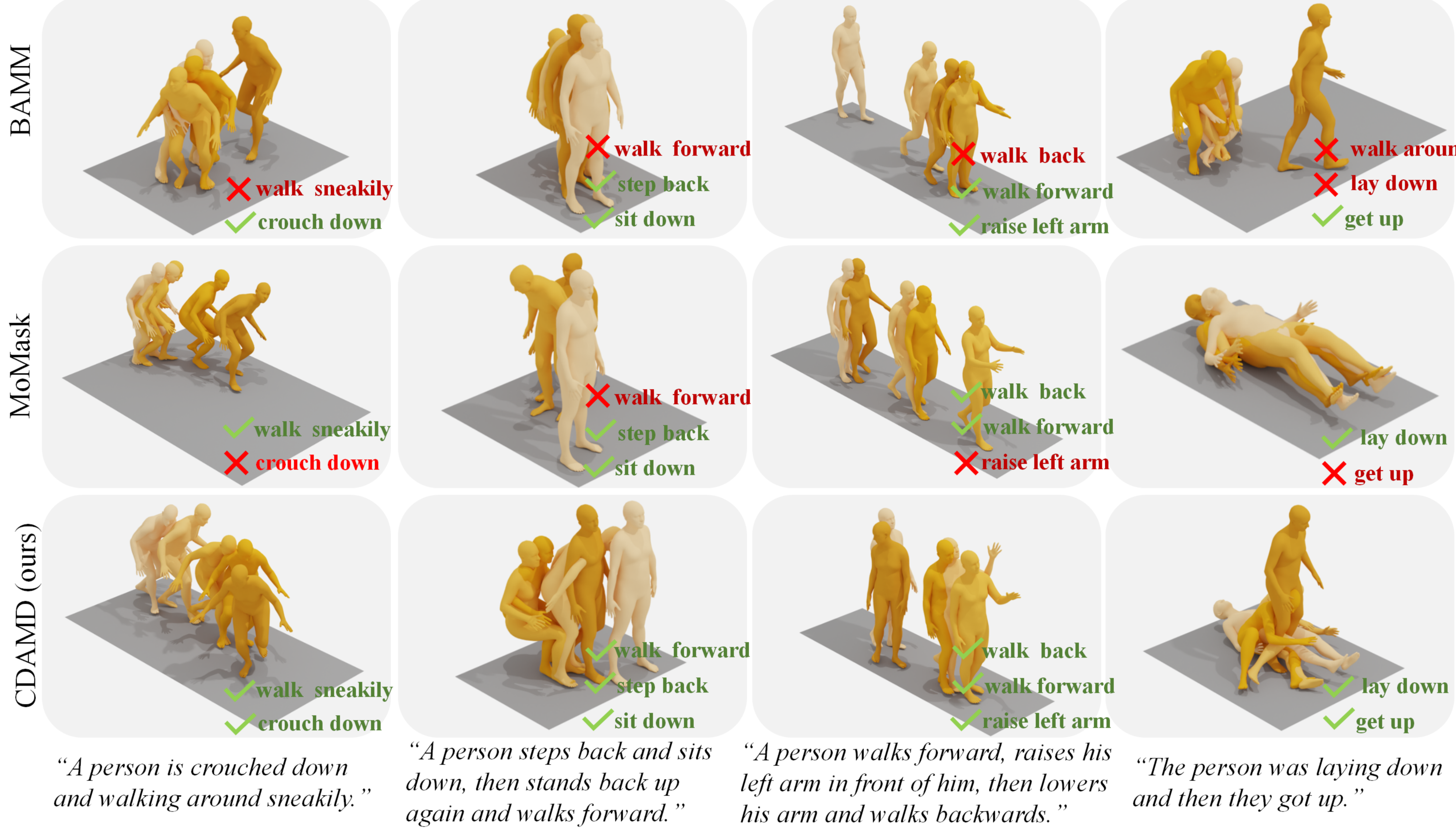}    
	\caption{Qualitative results of text-to-motion generation on HumanML3D. Our approach is compared with BAMM~\cite{pinyoanuntapong2024bamm} and MoMask~\cite{guo2024momask}, which are representative autoregressive motion models.}
	\label{fig:more_cp}
\end{figure*}

\begin{figure*}[htbp]    
	\centering    
	\includegraphics[width=\textwidth]{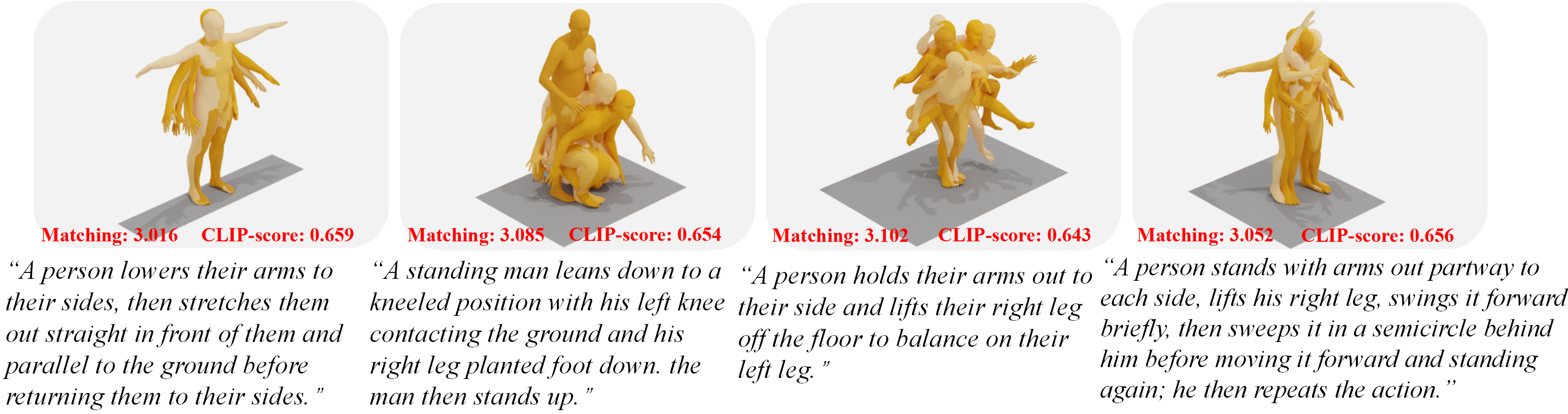}
	\centering\caption{The failure cases of text-to-motion on HumanML3D test set.}    
	\label{fig:failure_app_vs}  
\end{figure*}

\begin{figure}[tbp]    
	\centering    
	\includegraphics[width=0.46\textwidth]{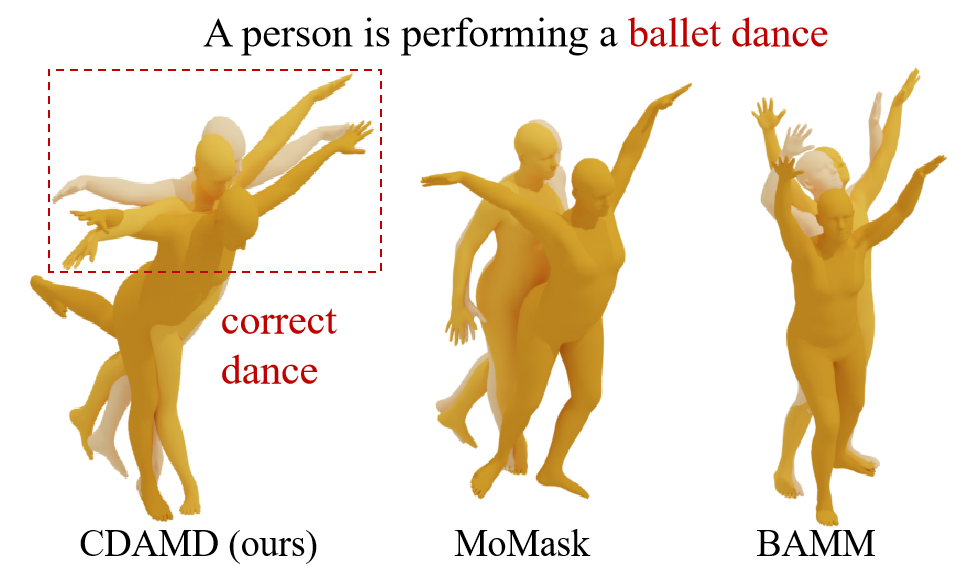}    
	\caption{Visualization comparison if textual to motion to state-of-the-art methods. Both the BAMM~\cite{pinyoanuntapong2024bamm} and MoMask~\cite{guo2024momask} generate motion of insufficient quality, while CDAMD generates higher quality and is more correlated with textual descriptions.}    
	\label{fig:cp}
\end{figure}

\subsubsection{Ablation Studies} 
Table \ref{tab:ablation study on HumanML3D} shows that removing either CMP or DCCA degrades performance under both coordinate-based and standard 263-dimensional mixed representations. In the coordinate-based setting, removing CMP increases FID from 0.082 to 0.136 and worsens Matching from 2.966 to 3.015, while removing DCCA causes a larger drop, increasing FID to 0.231 and Matching to 3.055. 
These results suggest that explicitly regulating temporal and conditional visibility is central to the effectiveness of CDAMD.

\subsubsection{Visualizations of Motion Editing} 
Fig.~\ref{fig:edit} demonstrates CDAMD's capability in various temporal editing tasks, including inpainting, outpainting, prefix, and suffix.

\subsubsection{Visualizations of Motion Generation}
We present additional qualitative comparisons of text-to-motion generation in Fig.~\ref{fig:more_cp}, 
highlighting the advantages of our approach over BAMM and MoMask on the HumanML3D dataset. 
As illustrated, our model consistently produces motion sequences that are better aligned with the text prompts 
and free from common artifacts such as incorrect action execution or missing temporal transitions. 

For example, given the prompt ``A person is crouched down and walking around sneakily.'',
our method faithfully generates a crouching posture followed by sneaky walking, 
while BAMM incorrectly emphasizes only walking and MoMask misinterprets the crouching action. 
These qualitative results demonstrate that our method achieves a higher degree of semantic fidelity 
and temporal precision compared to existing baselines. 
By effectively leveraging both textual cues and motion priors, 
our model generates motions that not only match the described actions but also preserve natural transitions 
between them.

\subsubsection{Analysis of Challenging Cases}
\label{analysis on failure cases}
We first analyze hard cases where our method fails. As illustrated in the Fig.~\ref{fig:failure_app_vs}, when given the prompt ``stretch them out straight in front of them,'' the generated motion keeps both arms positioned alongside the torso rather than extending them forward as instructed. In another example, with the prompt ``A person holds their arms out to their side and lifts their right leg off the floor to balance on their left leg,'' the generated sequence instead depicts the person alternating their supporting leg from left to right. These inaccuracies may stem from ambiguous or erroneous textual descriptions in the dataset, or they may indicate that our method still requires further refinement.

We then examine hard cases where our method produces correct results. Fig.~\ref{fig:cp} presents a qualitative comparison with MoMask~\cite{guo2024momask} and BAMM~\cite{pinyoanuntapong2024bamm} for text-to-motion generation. Notably, CDAMD generates motion accurately aligned with the provided text, whereas MoMask produces erroneous motion, and BAMM generates entirely inaccurate motion. 

\begin{figure}[tbp]            
	\centering            
	\includegraphics[width=0.46\textwidth]{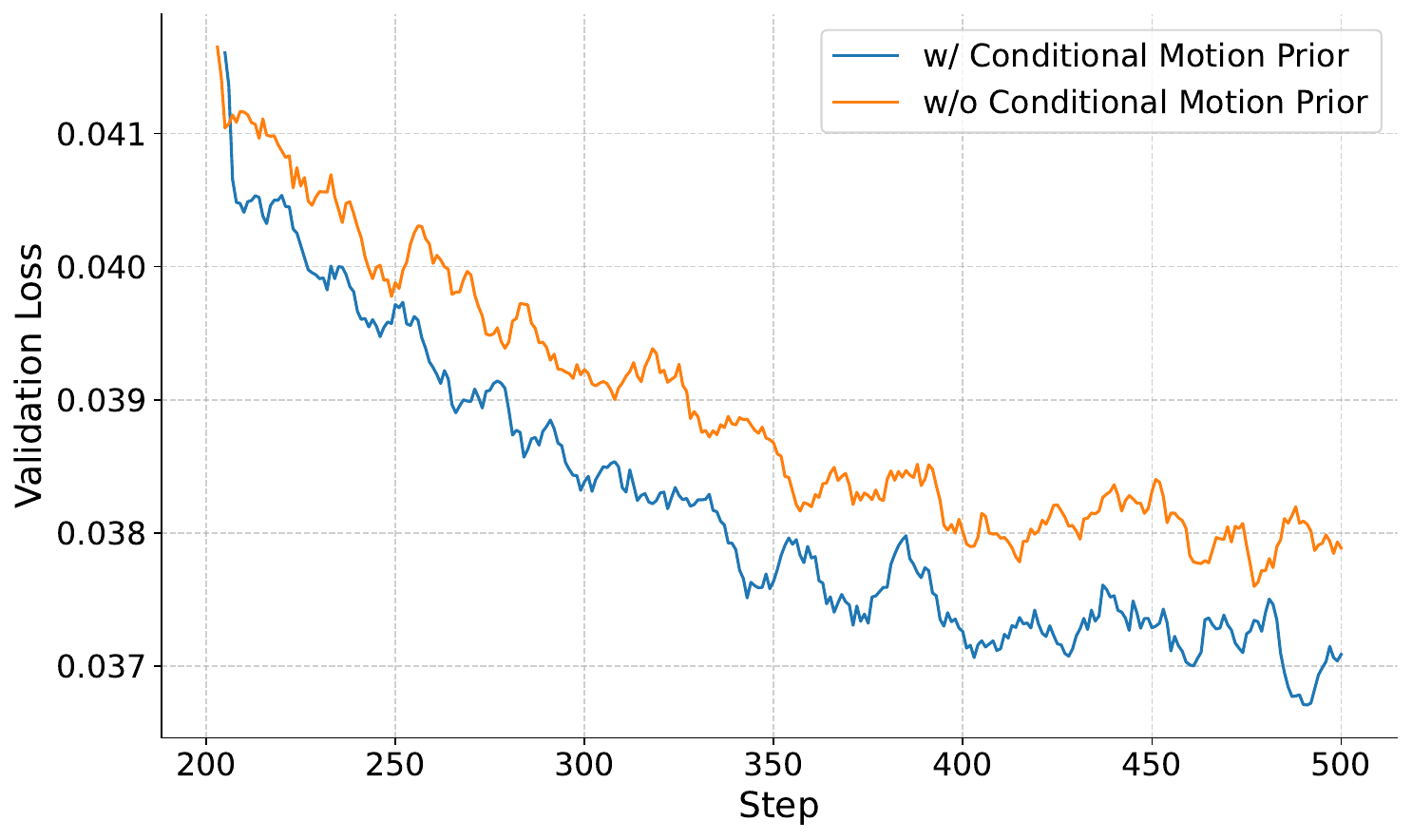}            
	\caption{Validation loss curves with and without conditional motion priors. Loss curves show faster convergence and lower error when motion priors are incorporated.}   
	\label{fig:val_loss}
\end{figure}

\begin{table}[t]
	\centering
	\caption{Comparison of motion preferences among CDAMD, MoMask~\cite{guo2024momask}, BAMM~\cite{pinyoanuntapong2024bamm}, and ground truth in the user study.} 
	\label{tab:motion_preference}  
	\resizebox{0.4\textwidth}{!}{ 
		\begin{tabular}{cccc} 
			\toprule
			CDAMD (ours) & MoMask & BAMM & Ground-Truth \\  
			\midrule
			39.8        & 21.5  & 10.4  & 28.3 \\  
			\bottomrule
		\end{tabular}
	}
\end{table}

\subsubsection{Impact of Motion Priors}  
As shown in Fig.~\ref{fig:val_loss}, the model with conditional motion priors achieves a lower validation loss curve, demonstrating not only improved generalization but also enhanced training efficiency. These results demonstrate that motion priors effectively guide the model toward more accurate and diverse motion generation.

\subsubsection{User Studies}
\label{user study}
We conduct user studies to assess generated motions with a particular focus on motion quality and naturalness. We recruit 20 participants, each of whom evaluates 15 sets of motion sequences rendered as videos. For each text prompt, participants are presented with three anonymized videos displayed side by side: one generated by our model (CDAMD) and two produced by strong baselines (MoMask and BAMM). Participants are asked to select the motion they find most realistic and visually appealing. As shown in Table \ref{tab:motion_preference}, motions generated by our model are preferred in 39.8\% of cases, substantially outperforming the baselines and achieving a quality level that closely approaches real motion capture data.

\subsubsection{Limitations} 
This iterative decoding achieves higher motion quality, but introduces longer inference latency. Another limitation is the dependency on the quality of the quantized motion tokens. 

\section{Conclusion}
This work presents CDAMD, a text-to-motion framework named Coordinate-based Dual-constrained Autoregressive Motion Generation.
CDAMD uses only coordinates as the motion representation and introduces the Conditional Motion Prior and Dual-Constrained Causal Mask for Transformer-based autoregressive motion generation. Ablation studies demonstrate the effectiveness of the proposed components. Since there is limited work on coordinate-based text-driven motion synthesis, we set up benchmarks for this setting and establish a variety of baseline approaches. Extensive experiments on motion generation and editing show that our approach produces flexible, high-fidelity, and semantically faithful motion generation.

\bibliographystyle{IEEEtran}
\bibliography{IEEEbib}

\end{document}